# Deep Learning Imputation of Missing Radius of Maximum Winds (Rmax) Values in Tropical Cyclone Best-Track Data

Swastik Agrawal[1,†], Nishkal Hundia[1,†], Ziyue Liu[2,3,*], Michelle Bensi[2]

[1]Department of Computer Science, University of Maryland, College Park, MD, USA

[2]Department of Civil and Environmental Engineering, University of Maryland, College Park, MD, USA

[3]Edwardson School of Industrial Engineering, Purdue University, West Lafayette, IN, USA

[†]These authors contributed equally to this work

[*]Corresponding author: ziyue20@terpmail.umd.edu

**Abstract**

Probabilistic coastal hazard assessments support risk-informed design of critical infrastructure and require characterization of the joint distribution of tropical cyclone (TC) parameters. However, TC datasets often contain missing records for the radius of maximum winds (Rmax), a key variable in Joint Probability Method analyses. Although conventional machine learning methods have been used to estimate missing Rmax, the potential of deep learning and physics-informed input augmentation remains insufficiently explored. This study evaluates data-driven strategies for Rmax imputation, including one-dimensional Convolutional Neural Networks (1DCNNs), Long Short-Term Memory (LSTM) networks, and conventional machine learning models. Physics-informed input augmentation and temporal modeling are examined, together with transfer learning using the synthetic RAFT and STORM datasets for pre-training and observational data from the International Best Track Archive for Climate Stewardship (IBTrACS) for fine-tuning. Results show that including the radius of 34-knot winds (R34) substantially improves performance across all model types. Temporal models achieve higher average correlations than non-temporal models despite being trained on approximately an order of magnitude fewer samples, indicating a stronger ability to preserve relative Rmax variability across storms. This advantage becomes more pronounced when R34 is unavailable, suggesting that temporal information can partially compensate for missing storm-size predictors. Transfer learning does not improve performance, likely because the synthetic datasets contain lower and less variable Rmax distributions than the observational IBTrACS data. These findings demonstrate the potential of temporal deep learning for reconstructing incomplete TC records and highlight the importance of physics-informed inputs, observational data availability, and distributional consistency for coastal hazard assessment.

**Plain Language Summary**

Historical tropical cyclone records are important for estimating coastal hazards. However, these records often have missing values for the radius of maximum winds (Rmax), which describes the distance from the storm center to the location of the strongest winds. Because Rmax is an important indicator of storm structure and size, missing values can reduce the accuracy of coastal hazard assessments. This study explores whether deep learning models can be used to estimate missing Rmax values in tropical cyclone best-track data. We compare temporal deep learning models, including one-dimensional convolutional neural networks and long short-term memory networks, with non-temporal machine learning models. We also test model performance under different model input variable sets. In addition, we investigate whether pre-training models on large synthetic datasets can improve performance when applied to observational data. The results show that temporal deep learning models can effectively impute missing Rmax values, especially when physics-informed input variables are used, but there are trade-offs that arise due to the relatively smaller size of datasets available for training. However, transfer learning from synthetic tropical cyclone datasets does not improve performance, likely because of differences in the Rmax data distributions between the synthetic datasets and the observational best-track data. These findings highlight both the promise of deep learning for improving historical storm records and the importance of matching synthetic training data to real observational distributions.

## 1. Introduction

Storm-induced coastal hazards can impose a variety of adverse loads and conditions on infrastructure systems. The probabilistic hazard assessments of coastal hazards often rely on historical storm data (e.g., Nadal-Caraballo et al. 2022; Liu et al. 2024a, 2025a, 2025b; Liu and Bensi 2026; Cheng et al. 2026). However, historical storm records are often temporally or spatially incomplete, particularly in the earlier years of record (Liu et al. 2024b). One particularly important yet often missing storm parameter is the radius of maximum winds ($R_{max}$), which is a representation of the size of a storm. Due to its physical characteristics, $R_{max}$ plays a critical role in the generation of multiple coastal hazards, and it is necessary to characterize its distribution as part of probabilistic hazard assessments (Chavas et al. 2015; Chavas and Knaff 2022). Thus, it is important to accurately estimate the value of $R_{max}$ in datasets where it is missing.

The ultimate goal of this study is to predict missing $R_{max}$ values in global historical tropical cyclone (TC) datasets by developing and comparing multiple deep learning approaches. Figure 1 shows an example of the imputation process. Given a storm track containing gaps in its observed $R_{max}$ record (highlighted by red squares), we aim to impute these missing values to return a fully reconstructed $R_{max}$ time history.

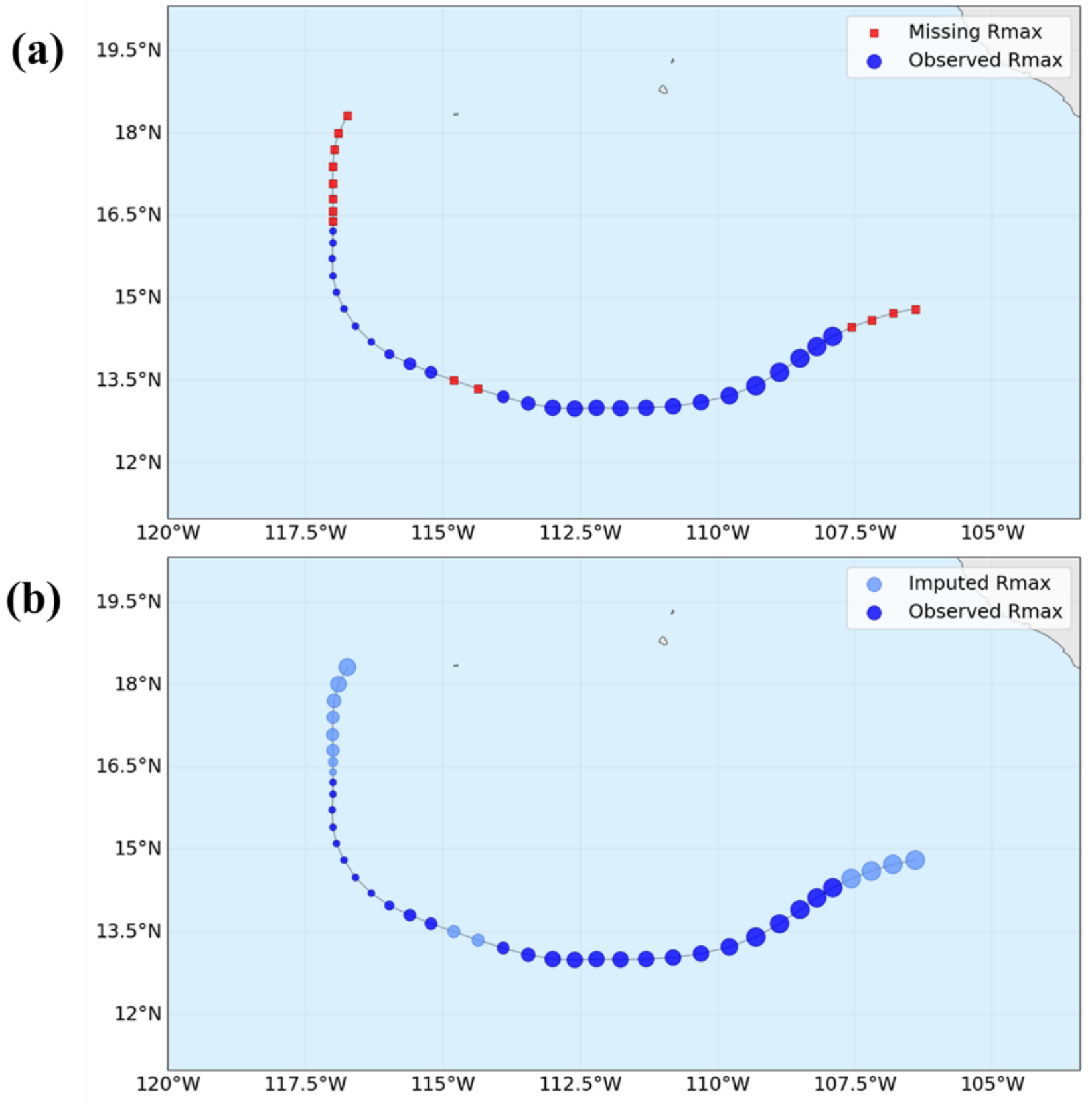

Figure 1. Illustration of $R_{max}$ imputation along a storm track: (a) available observations, shown as blue circles scaled by $R_{max}$, and missing values, shown as red squares; and (b) the reconstructed track after the missing $R_{max}$ values are imputed.

### 1.1 Classical methods for estimating $R_{max}$

Classical approaches to estimating $R_{max}$ are primarily based on physics-derived or empirical relationships. Vickery and Wadhera (2008) proposed a regression-based formulation that incorporates latitude and pressure deficit for Atlantic storms. Chavas et al. (2015) developed physically based models for the radial structure of tropical cyclones, linking $R_{max}$ to the overall wind structure through angular momentum conservation. This framework was further refined by Chavas and Knaff (2022), who emphasized the dependence of $R_{max}$ on storm size and structure. In particular, they used the radius of 34-kt winds ($R_{34}$) as a key predictor in a regression-based model for estimating $R_{max}$, demonstrating strong model performance. Empirical formulations have also been proposed that estimate $R_{max}$ directly from the radius of 50-kt winds ($R_{50}$). Using typhoon observations and best-track data over the western North Pacific, Takagi and Wu (2016) derived a simple relationship $R_{max} = 0.23R_{50}$, and demonstrated that accounting for variability in $R_{max}$ is critical for accurate storm surge simulations. While these methods provide physically interpretable estimates, their applicability can be limited across basins, storm intensities, and climatological regimes. Additionally, in some cases the input parameters for these methods are less frequently available than $R_{max}$ itself, limiting their applicability in imputing $R_{max}$.

### 1.2 Machine learning for $R_{max}$ imputation

ML techniques have been successfully applied to a range of TC hazard modeling tasks (e.g., Jia and Taflanidis 2013; Al Kajbaf and Bensi 2020; Lee et al. 2021; Liu et al. 2024c, 2025c). However, they have seen limited application in $R_{max}$ estimation. Liu et al. (2024b) used ML methods to impute missing $R_{max}$ values in best-track datasets. Their results demonstrated improved robustness and reduced bias compared to classical approaches. Nevertheless, these models typically treat individual storm observations independently and are limited in their ability to capture the temporal dependencies inherent in storm evolution.

### 1.3 Deep learning for Tropical Cyclone hazards and $R_{max}$ imputation

Deep learning models provide a natural extension of traditional ML by enabling the representation of complex nonlinear and temporal relationships in storm evolution. Architectures such as Long Short-Term Memory (LSTM) networks have demonstrated strong performance in TC track and intensity prediction (Tong et al. 2022), while one-dimensional convolutional neural networks (1DCNNs) have been applied to TC-related time series tasks such as peak storm surge prediction (Lee et al. 2021). More broadly, convolutional neural networks (CNNs), LSTMs, and hybrid CNN-LSTM architectures have been shown to outperform traditional machine learning methods for storm surge forecasting and other coastal hazard applications (Tiggeloven et al. 2020; Wang et al. 2020). Recent studies have further explored advanced architectures, including bidirectional LSTMs, transformers, and attention-based networks, achieving additional improvements in TC prediction skill (Mulia et al. 2025).

Despite these successes, the application of deep learning to the imputation of missing $R_{max}$ values in historical best-track datasets has not been widely implemented. Xu et al. (2024a) employed a CNN to reconstruct $R_{max}$ from ERA5-derived gridded wind profile input data, demonstrating that deep learning can reduce systematic biases in reanalysis-based storm size estimates. However, deep learning methods have not been implemented to address the data imputation problem within incomplete storm observational records.

### 1.4 Study objective and contributions

In this study, we compare several approaches to impute missing $R_{max}$ values within the global best-track dataset. We consider several model input configurations, including a physics-augmented configuration. We evaluate temporal deep learning architectures that preserve storm-level temporal structure and compare the performance trade-offs of considering temporal architectures through comparison against pointwise prediction approaches. We further investigate transfer learning strategies using synthetic TC datasets (RAFT (Xu et al. 2024b) and STORM (Bloemendaal et al. 2020)) to mitigate data scarcity and improve generalization.

## 2 Data and Methods

This section presents the data and methods workflow of the study. Figure 2 provides an overview of the analysis procedure, from observational data processing to model development, performance evaluation, and transfer learning analysis.

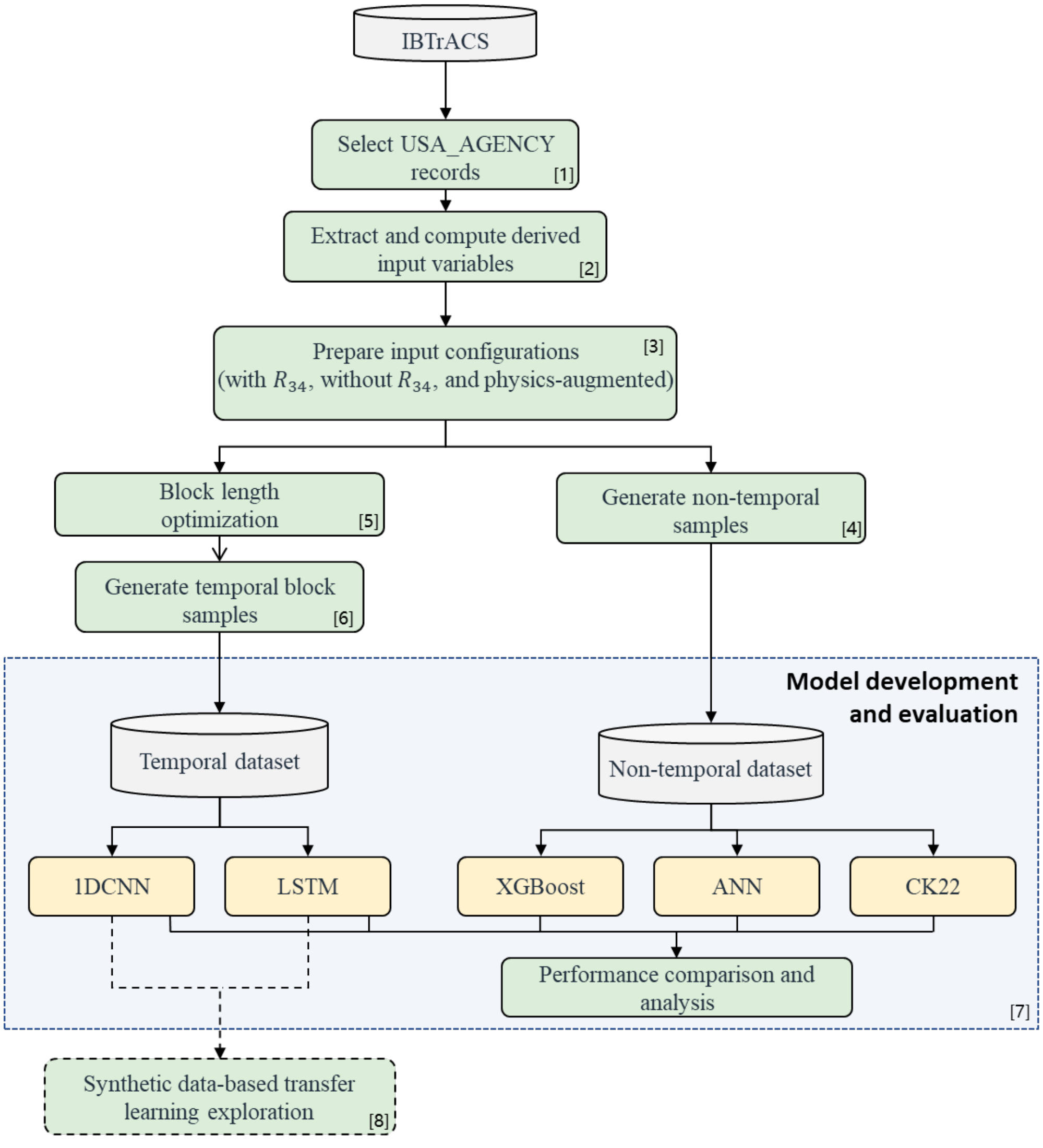

Figure 2. Overview of the data processing, model development, and evaluation workflow.

## 2.1 Observational data sources

We use historical TC reanalysis data sourced from the International Best Track Archive for Climate Stewardship (IBTrACS) as the observational data source (Demuth et al. 2006; NOAA 2021). The IBTrACS compiles initial storm observations and subsequent reanalysis from sources such as reports from ships, land stations, radars, satellites, etc. We use the USA_AGENCY data series in IBTrACS, covering records from 1842 to 2023, because it provides the most consistent spatial and temporal coverage and broad availability across variables. The temporal resolution of the data is generally 3 hours, and the spatial resolution is 0.1 degrees (NOAA 2021). The availability of $R_{max}$ records in IBTrACS across all basins from the USA_AGENCY (which also provides the most complete $R_{max}$ records in the IBTrACS dataset) is shown in Figure 3.

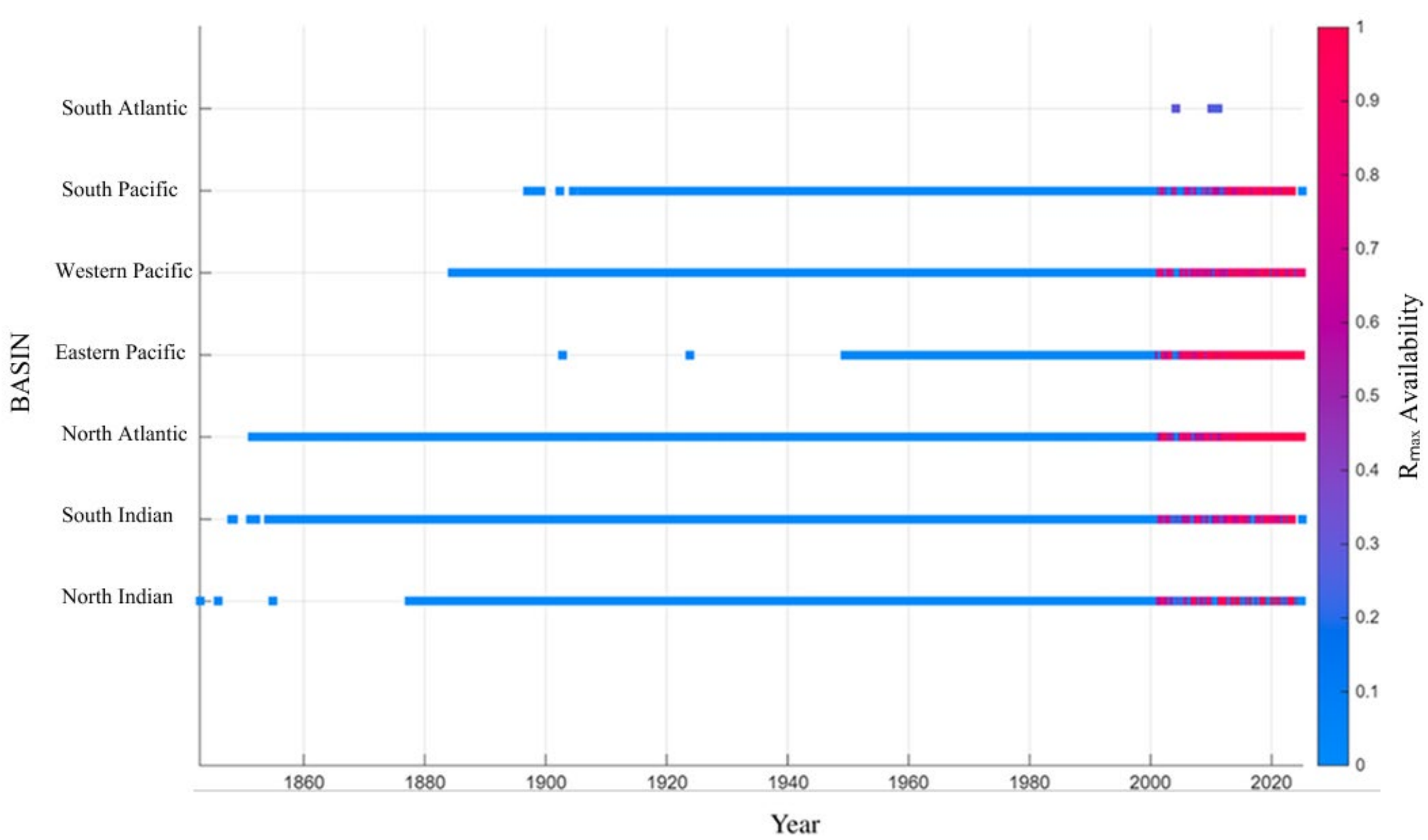

Figure 3. $R_{max}$ reanalysis data availability (values represent the ratio of $R_{max}$ recorded in each single storm) in each basin in IBTrACS dataset USA_AGENCY data. White portions in each bar indicate periods with no historical storm data recorded.

## 2.2 Data processing

As illustrated in Figure 2, Box [1], we begin by filtering the input dataset to include only USA_AGENCY data, ensuring consistency in the extracted data and measurement standards. We retain only the variables relevant to our model development, as given in Table 1. These variables were chosen based on previous work imputing $R_{max}$ using ML (Liu et al. 2024b).

Table 1. Extracted input/output variables from IBTrACS

| Variable | Symbol | Unit |
|---|---|---|
| Radius of maximum winds | $R_{max}$ | nmile (nautical miles) |
| Minimum pressure at sea level | $P_c$ | hPa |
| Maximum sustained wind speed | $V_{max}$ | knots |
| Longitude | $lon$ | degrees, [-180, 180] |
| Latitude | $lat$ | degrees, [-90, 90] |
| Wind radii of 34 knots (mean of non-zero quadrants) | $R_{34}$ | nmile (nautical miles) |
| Distance of the storm point to land | $d$ | nmile (nautical miles) |
| Storm forward speed | $V_f$ | knots |
| Storm heading direction | $\theta$ | degrees [0,360], clockwise with due north as 0 degrees |

We next compute several derived input variables (see Figure 2, Box [2]). We introduce a categorical flag ($IS_{land}$) in our input variables indicating whether the storm data point is over land or water. This flag is obtained by identifying whether each storm data point's coordinates fall over land or water, using a global reference grid that labels every location (~1km resolution) on Earth (Karin 2020). To account for the circular nature of storm heading, we represent the direction angle using its sine and cosine components and include both as derived input variables. To improve the representation of spatial information in the neural network models, we transform latitude and longitude into three-dimensional Earth-centered, Earth-fixed (ECEF) Cartesian coordinates (X, Y, Z) (UNOOSA 2012). This transformation avoids the discontinuities and distance distortions associated with raw latitude-longitude coordinates and provides a continuous Euclidean representation of storm location. The equations used to compute the ECEF coordinates are provided in Appendix A. Table 2 summarizes the derived input variables.

Table 2. Derived input variables

| Variable | Symbol(s) | Unit |
|---|---|---|
| Land indicator flag (binary) | $IS_{land}$ | N/A |
| Sine of storm heading direction | $\sin(\theta)$ | N/A |
| Cosine of storm heading direction | $\cos(\theta)$ | N/A |
| ECEF X, Y, Z-coordinates | $X, Y, Z$ | km |

## 2.3 Input configuration

We experiment with three different ML model input configurations, as shown in Table 3 and reflected in the overall modeling framework as Figure 2, Box [3]. In Table 3, the first column indicates the naming convention that will be used in the study to refer to each configuration, and the second column provides the constituent input variables.

Table 3. Input configurations

| Configuration | Input Variables | Output Variable |
|---|---|---|
| With $R_{34}$ | $P_c, V_{max}, R_{34}, d, V_f, IS_{land}, \sin(\theta), \cos(\theta), X, Y, Z$ | $R_{\max}$ |
| Without $R_{34}$ | $P_c, V_{max}, d, V_f, IS_{land}, \sin(\theta), \cos(\theta), X, Y, Z$ | $R_{\max}$ |
| Physics-augmented | $P_c, V_{max}, R_{34}, d, V_f, IS_{land}, \sin(\theta), \cos(\theta), X, Y, Z, R_{34}^2, \sin(\text{lat}), f, \frac{M_{max}}{M_{34}}$ | $R_{\max}$ |

The first two input configurations are the same except for the inclusion of $R_{34}$. We test both the inclusion and exclusion of $R_{34}$ because $R_{34}$ is known to be highly correlated with $R_{max}$ but is not widely available in historical datasets (Liu et al. 2024b). The physics-augmented input configuration is based on Chavas and Knaff (2022) and introduces additional parameters derived from empirical relationships. The hypothesis is that incorporating these known physical relationships can reduce the number of training steps required for the model compared to when the model must learn them independently. This effectively embeds established TC physics into the dataset, potentially enhancing model performance.

The additional input variables in the physics-augmented input configuration were derived from the equations (2), (3), (4), and (7) in Chavas and Knaff (2022) after accounting for the differences in units. These include $R_{34}^2$, $f$, sin(lat), and the ratio $M_{\max}/M_{34}$, where $f$ is the Coriolis parameter, $M_{\max}$ is the absolute angular momentum at $R_{\max}$, and $M_{34}$ is the absolute angular momentum at $R_{34}$. The terms are defined as:

$$f \quad = \quad 2\,\Omega \sin(\text{lat}) \tag{1}$$

$$M_{34} = R_{34} \times 17.5ms^{-1} + 0.5fR_{34}^2 \tag{2}$$

$$\frac{M_{max}}{M_{34}} = 0.699\,\exp[-0.00618(V_{max} - 17.5ms^{-1}) - 0.00210(V_{max} - 17.5ms^{-1})0.5fR_{34}^2] \tag{3}$$

where $\Omega = 7.292 \times 10^{-5}\ \text{s}^{-1}$ is Earth's rotation rate.

To enable the use of non-temporal models, all available data are pooled into a single non-temporal dataset (see Figure 2, Box [4]). Even though data is pooled, storm indices are retained to support later storm-based cross-validation. This allows us to enforce storm-level data separation for training, testing, and validation. To enable the use of temporal models, the workflow proceeds to the specification of temporal blocks, as shown in Figure 2, Boxes [5]-[6], and described in the next section.

### 2.4 Block length optimization and generation of block samples

We structure the time series data for input into sequential models such as CNNs and LSTMs by dividing each storm's track into fixed-length segments, referred to herein as blocks. Each block represents a contiguous subsequence of the storm's time series, capturing the evolution of relevant variables over a fixed window of time.

Blocks are generated separately for each storm and are temporally homogeneous. Typically, though not exclusively, there exists a consistent 3-hour time interval between two consecutive data points reported in IBTrACS. As with the pooled non-temporal data, we store metadata about each block's origin (i.e., storm ID as well as start and end indices of each block in the original dataset) to support later storm-based cross-validation.

To identify the optimal block length, we perform a preliminary sensitivity study in which we train 1DCNN and LSTM models using a four-fold cross-validation split across the entire dataset. The workflow of this block length optimization is provided in Figure 4.

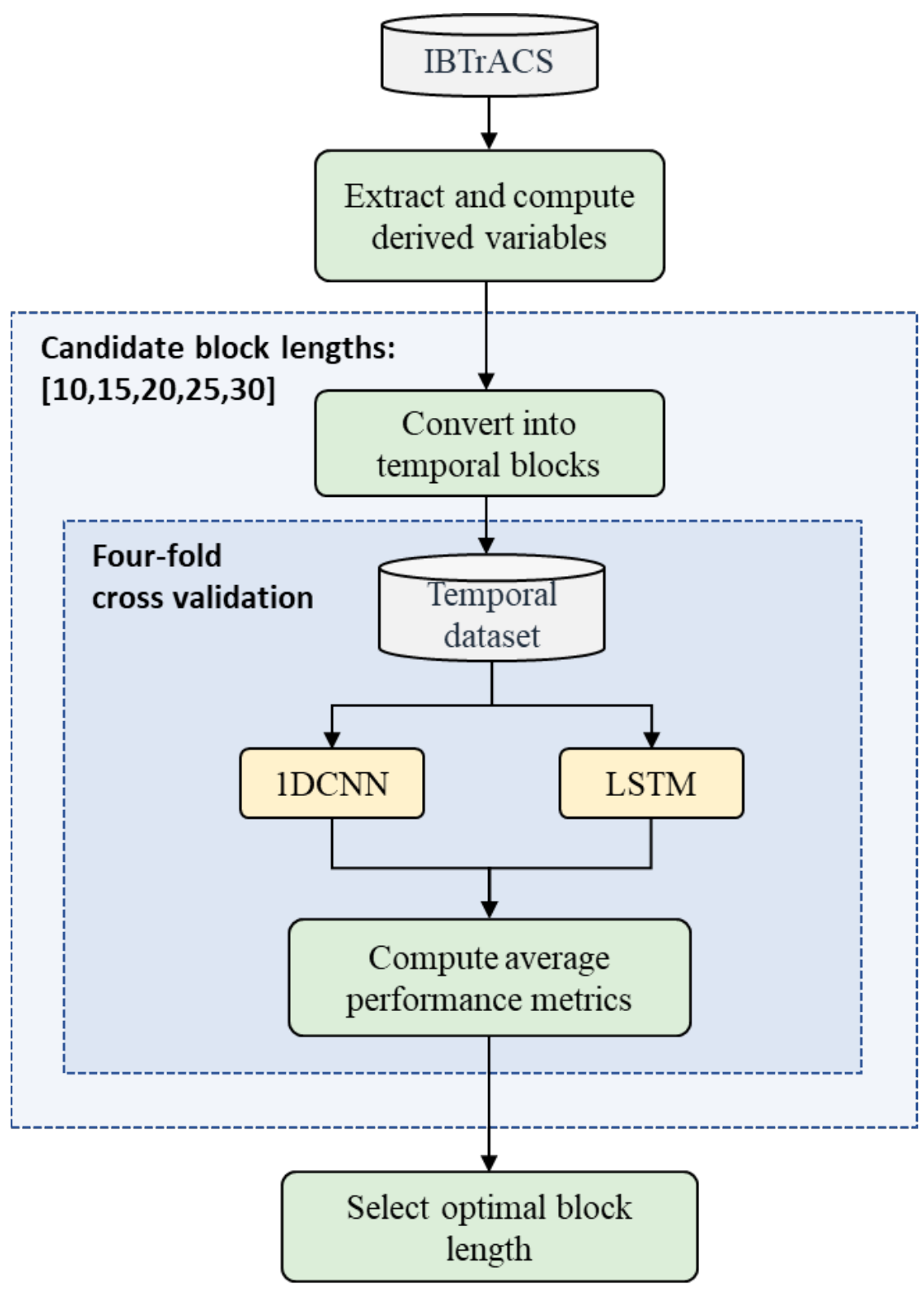

Figure 4. Workflow of block length optimization

For each fold, blocks are partitioned into training and testing sets, maintaining strict isolation between storms across these sets. That is, all blocks from a given storm are assigned exclusively to one of two sets in each fold and are never split across them. Multiple fixed lengths for the blocks were tested (10, 15, 20, 25, and 30 time steps). The average Pearson correlation coefficients ($R$) between observed and predicted $R_{max}$ values across cross-validation folds for each block length and model type are reported in Figure 5.

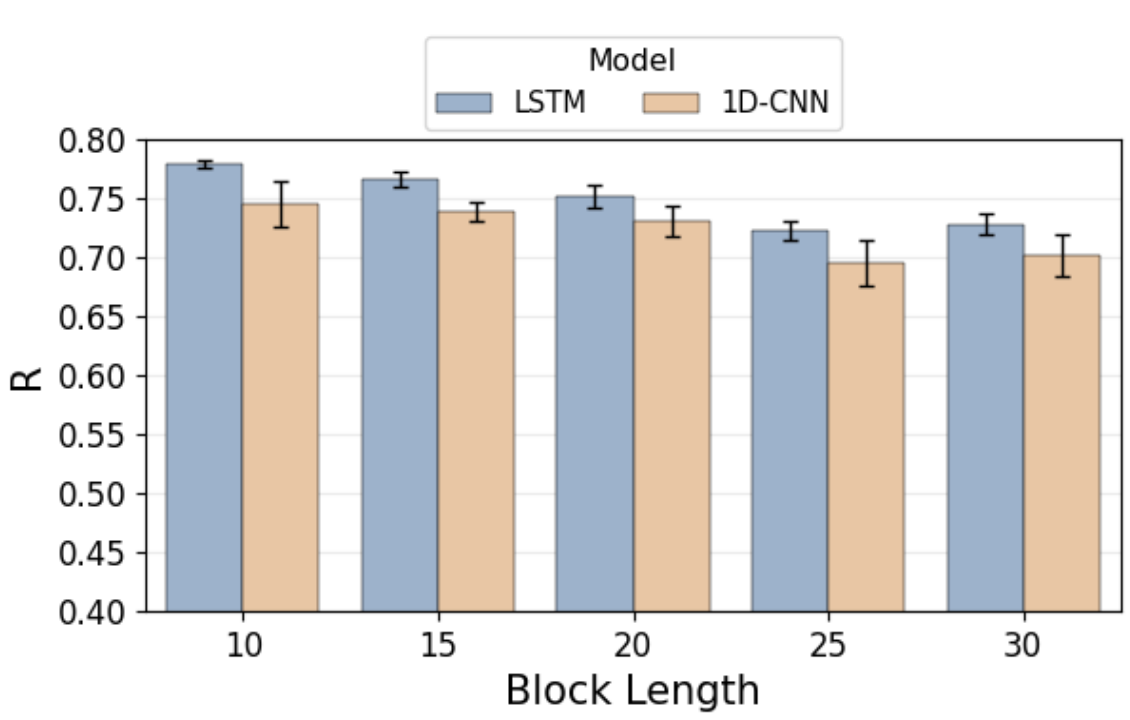

Figure 5. Averaged Pearson correlation coefficients between observed and predicted $R_{max}$ values of various block lengths. Error bars indicate the range (min, max) across folds.

As is evident in Figure 5, a block length of 10 yields the highest correlation coefficient between observed and predicted $R_{max}$ values. We observe a gradual reduction in model performance as block length increases. This is primarily due to the reduction in the number of usable training samples. Blocks require contiguous time steps, and longer contiguous sequences become less available as block length increases, resulting in fewer constructed blocks. Furthermore, since many storms contain gaps in their $R_{max}$ observational records, longer blocks are more likely to include at least one missing $R_{max}$ value, resulting in a decrease in the number of valid training samples. Other evaluation metrics, including Mean Absolute Error (MAE) and Mean Squared Error (MSE), corroborate this finding. Thus, we conclude that 10 is the most effective block length.

### 2.5 Model development and evaluation

We evaluate deep learning models, specifically 1DCNNs and LSTMs, because of their ability to capture temporal dependencies in storm evolution. The deep learning models are compared against non-temporal ML models, namely XGBoost (eXtreme Gradient Boosting) and ANN (artificial neural network). The model development and evaluation portion of the workflow is reflected in Figure 2, Box [7].

1DCNNs operate by sliding convolutional filters over the input sequence to detect local temporal patterns. These models have been shown to be effective in predicting short-term dependencies and trends in TC storm surge time series data (Lee et al. 2021). This architecture consists of convolutional layers followed by pooling layers, which reduce dimensionality as information passes through the model.

LSTMs, on the other hand, are designed to model long-term dependencies in sequential inputs using gated memory cells (Hochreiter and Schmidhuber 1997). This makes them suitable candidates for capturing the progression of storm characteristics over time (Gao et al. 2018). By maintaining both long-term and short-term memory, LSTMs are capable of retaining localized information about storm points near the current timestep, while also preserving broader contextual information about the overall storm track.

To evaluate the effectiveness of temporal deep learning models, we compare the performance of the temporal deep learning models (1DCNN and LSTM) against the two non-temporal ML models: XGBoost and ANN. These non-temporal ML models are limited in their ability to capture the temporal dependencies and evolving dynamics within a storm's lifecycle. However, because they operate on pointwise observations and do not require sequential time blocks, substantially more data is available for training.

XGBoost helps capture complex nonlinear relationships and interactions within the data by sequentially constructing an ensemble of decision trees, where each tree corrects residuals via regularized gradient descent (Chen and Guestrin 2016). The ANN, on the other hand, approximate nonlinear input-output relationships through stacked layers of learned weights. Additionally, we benchmark against the empirical formula outlined in Chavas and Knaff (2022), herein referred to as CK22, which serves as a physics-informed regression model for comparison against our data-driven approaches.

The overall model development and evaluation workflow is summarized in Figure 6. The process begins with architecture selection and hyperparameter tuning. For 1DCNN and LSTM models, we perform a random search with 100 trials for each model over hyperparameters such as the number of layers, number of neurons, and activation functions (Bergstra and Bengio 2012). For XGBoost, the hyperparameters were selected using a coarse grid search over predefined candidate values for learning rate, maximum tree depth, and number of estimators. The ANN architecture used in this study is adopted from Liu et al. (2024b). The neural network architectures employed in this study, including 1DCNN, LSTM, and ANN, are presented in Figure 7.

Model training and performance assessment is performed using four-fold cross-validation in which storms are assigned exclusively to either the testing set or the training set in each fold and are never split across them. We apply min-max scaling to all variables except $\sin(\theta)$, $\cos(\theta)$, and $IS_{land}$ :

$$x_{\text{scaled}} = \frac{x - x_{\min}}{x_{\max} - x_{\min}} \tag{4}$$

To prevent information leakage, the scaling parameters are estimated separately within each training set and then applied to the corresponding testing set.

Models are trained and tested using the normalized data set. A reverse normalization is applied before the performance assessment so that performance metrics are computed in the physical units. Due to data scarcity, the same dataset is used for architecture selection, hyperparameter optimization, and model training. This reuse may introduce limited data leakage in the model performance comparison.

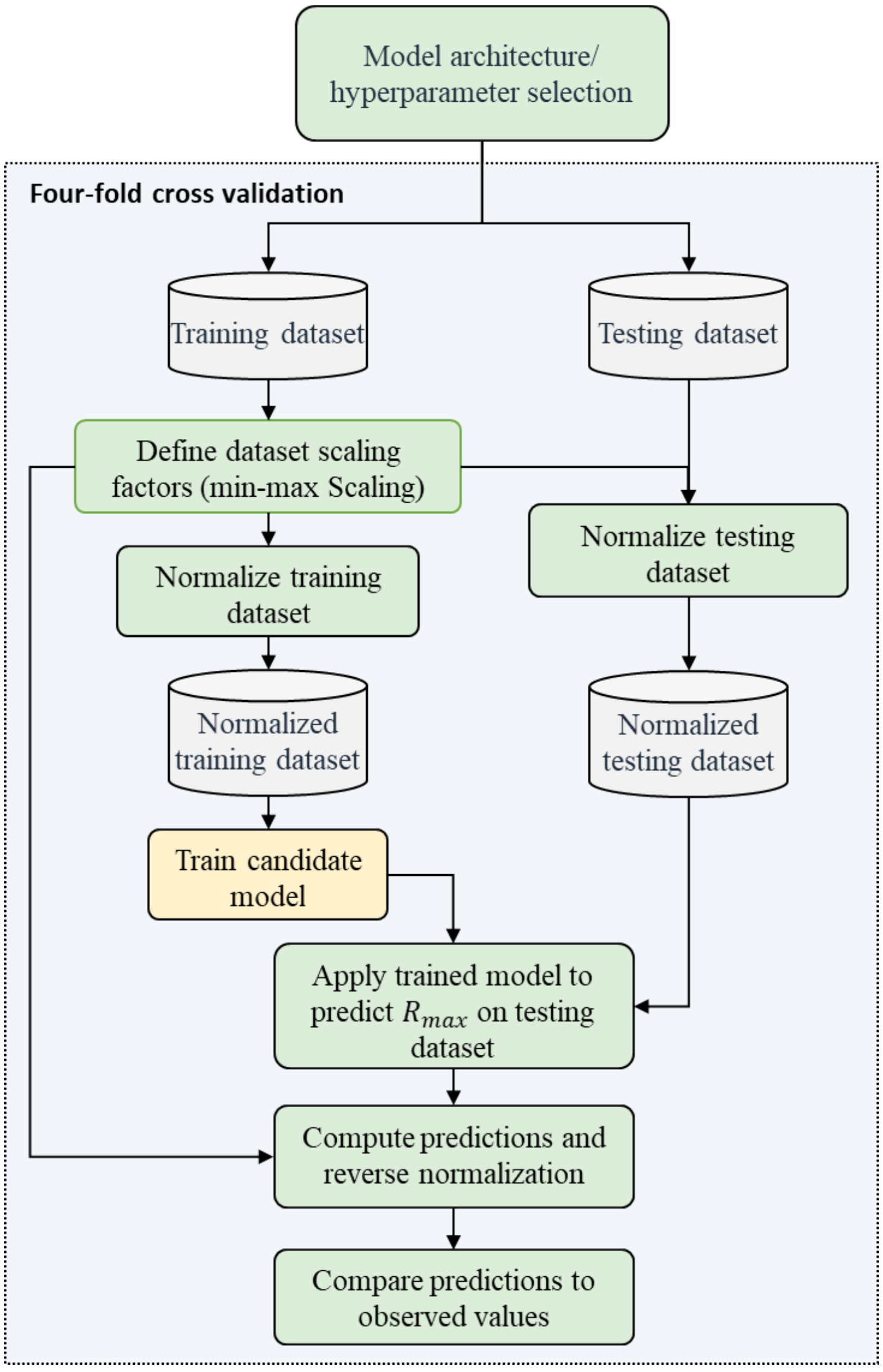

Figure 6. Workflow of model development and evaluation

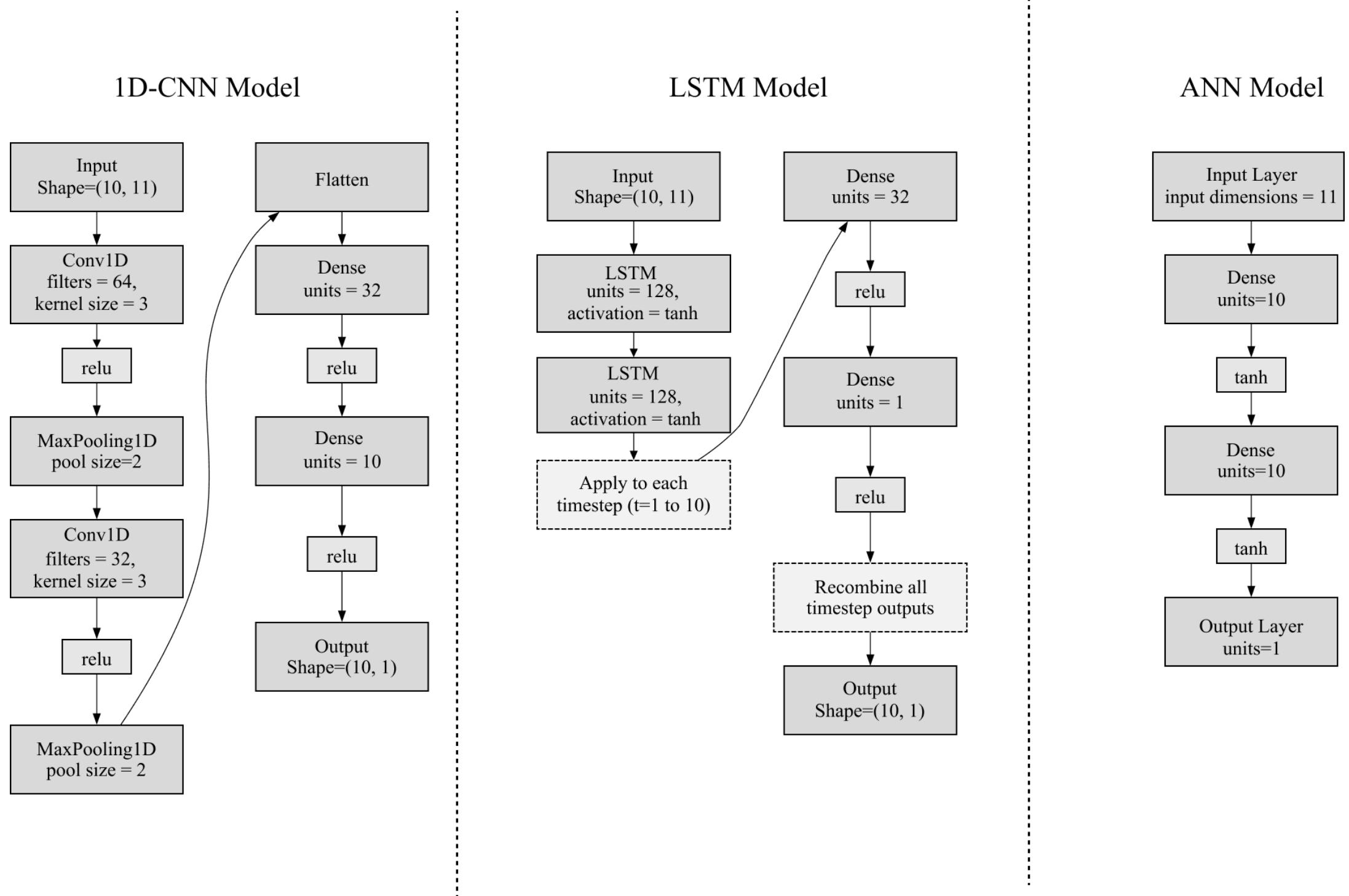

Figure 7. Flowchart depicting model architecture for the tested neural networks.

We also experimented with Gaussian Process Regression (GPR) using the configuration identified as most effective in Liu et al. (2024b), training the model on a random subsample of 2,000 data points from the training set. However, we exclude GPR from the final comparison because its training complexity scales cubically with the number of training samples. Figure 8 illustrates this limitation by comparing total training floating-point operations (FLOPs), a hardware-agnostic measure of computational cost, for GPR, LSTM, and 1DCNN models as a function of dataset size. The cubic scaling associated with GPR kernel matrix inversion rapidly leads to substantially higher computational cost than the other methods. While GPR achieves comparable performance when trained on 2,000 points, its cubic scaling in training complexity renders it impractical for larger datasets, which is why it has been excluded from the results presented in the remainder of this paper.

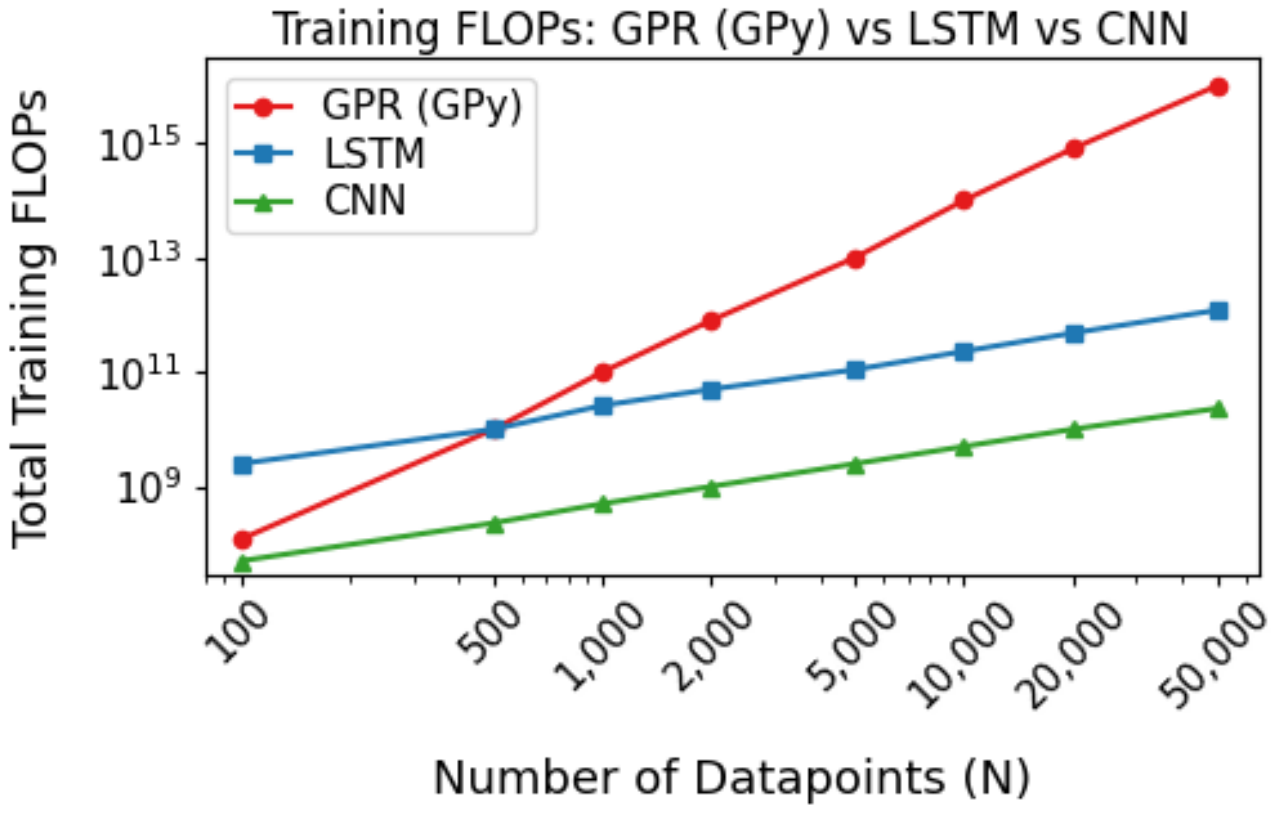

Figure 8. Comparison of total training FLOPs across GPR (GPy), LSTM, and 1DCNN as a function of training set size on a log-log scale.

### 2.6 Synthetic data-based transfer learning

To address the data scarcity challenges that can affect the performance of temporal deep learning models, we explored a transfer learning approach using synthetically generated TC datasets (Figure 2, Box [8]). The underlying hypothesis was that pre-training models on larger synthetic datasets would enable them to learn broader correlations between storm parameters, followed by fine-tuning on IBTrACS data to capture real-world interactions and edge cases that may not be adequately represented in synthetic data. We employed two synthetic datasets for this investigation:

- **RAFT (Risk Assessment of coastal Flooding from Tropical cyclones):** The RAFT dataset (Xu et al. 2024a) simulates 40,000 synthetic TCs for the North Atlantic using a hybrid framework combining statistical methods, physical models, and deep learning. It provides high-resolution synthetic storm tracks, intensities, and rainfall patterns at 6-hour intervals, capturing key risk factors like wind, precipitation, and landfall behavior.
- **STORM (Synthetic Tropical cyclOne geneRation Model):** The STORM dataset (Bloemendaal et al. 2020) uses a fully statistical resampling model to extend 38 years of historical cyclone data into 10,000 years of global TC activity. It generates storm tracks, intensities, and sizes across all major basins.

Both datasets include the same key parameters as IBTrACS ($P_c$, $V_{max}$, $R_{max}$, latitude, longitude, etc.), and provide substantially larger sample sizes than IBTrACS, potentially allowing temporal models to better learn underlying storm dynamics before exposure to real-world data complexities.

The transfer learning methodology involved first pre-training the 1DCNN and LSTM architectures on the synthetic datasets using the same block-based approach described in Section 2.4. The pre-trained models were then fine-tuned on IBTrACS data, with the expectation that this approach would leverage the benefits of both synthetic data abundance and real-world data fidelity.

## 3 Results and Discussion

### 3.1 Model performance analysis

To evaluate model performance, we use two primary metrics: the correlation coefficient ($R$) and the root mean squared error (RMSE) between predicted and observed $R_{max}$ values. These metrics provide complementary insights into each model's ability to impute missing $R_{max}$ values in storm data sequences.

Specifically, $R$ provides insights into each model's ability to capture variation and patterns across storms. RMSE captures the average magnitude of prediction errors and provides insights regarding overall prediction accuracy. We compare the deep learning models (1DCNN and LSTM) against the non-temporal baselines (XGBoost and ANN). Performance metrics are computed across four cross-validation folds, and Table 4 summarizes the model performance across all folds.

It is important to note that temporal deep learning models (LSTM and 1DCNN) are trained on significantly fewer samples than non-temporal models such as XGBoost and ANN. This is due to the need to construct temporally coherent input sequences (e.g., 10 time steps per sample), which limits the number of usable blocks in the dataset. Table 4 includes the sizes of the training and testing sets, which differ by an order of magnitude between temporal and non-temporal models.

Table 4. Model performance comparison. All reported metrics are presented as mean [min, max] across five random seeds; metrics were first averaged across the four cross-validation folds within each seed.

| Model | *R* value | RMSE (nm) | Train Samples* | Test Samples* |
|---|---|---|---|---|
| **Temporal models** | | | | |
| 1DCNN with $R_{34}$ | 0.770 [0.765, 0.775] | 12.91 [12.75, 13.07] | 3,300 | 1,100 |
| *1DCNN without $R_{34}$* | *0.616 [0.611, 0.621]* | *17.35 [17.21, 17.49]* | *5,500* | *1,800* |
| 1DCNN with physics-augmented | 0.771 [0.765, 0.777] | 12.89 [12.77, 13.01] | 3,300 | 1,100 |
| LSTM with $R_{34}$ | 0.779 [0.777, 0.781] | 12.70 [12.53, 12.87] | 3,300 | 1,100 |
| *LSTM without $R_{34}$* | *0.620 [0.617, 0.623]* | *17.28 [17.16, 17.40]* | *5,500* | *1,800* |
| LSTM with physics-augmented | 0.781 [0.777, 0.785] | 12.66 [12.44, 12.88] | 3,300 | 1,100 |
| **Non-temporal models** | | | | |
| XGBoost with $R_{34}$ | 0.760 [0.758, 0.762] | 13.31 [13.27, 13.35] | 45,000 | 14,400 |
| *XGBoost without $R_{34}$* | *0.570 [0.566, 0.574]* | *18.28 [18.21, 18.35]* | *70,900* | *23,500* |
| XGBoost with physics-augmented | 0.759 [0.758, 0.760] | 13.31 [13.27, 13.35] | 44,000 | 15,400 |
| ANN with $R_{34}$ | 0.658 [0.640, 0.676] | 15.45 [15.08, 15.82] | 44,600 | 14,800 |
| *ANN without $R_{34}$* | *0.536 [0.525, 0.547]* | *18.70 [18.43, 18.97]* | *71,200* | *23,200* |
| ANN with physics-augmented | 0.669 [0.664, 0.674] | 15.38 [15.24, 15.52] | 44,700 | 14,800 |
| CK22 with $R_{34}$ | 0.705 [0.704, 0.706] | 17.72 [17.71, 17.73] | — | 61,200† |

*The sample sizes correspond to one representative fold and vary slightly across folds.
†Because CK22 is an existing parametric model fitted in an independent study, no training within this study is required, and all data points are used for testing.

Overall, the performance metric results indicate that the direct inclusion of $R_{34}$, as well as the use of the physics-augmented input configurations, results in notably better model performance than configurations excluding $R_{34}$. This improvement is evident in both correlation and RMSE. Among models leveraging $R_{34}$, the temporal models consistently achieved higher average correlation than the non-temporal models evaluated in this study, despite being trained on datasets approximately an order of magnitude smaller. The physics-augmented LSTM also achieved the lowest average RMSE (12.66 nm), although RMSE differences among the best-performing $R_{34}$-based models were relatively small. These results suggest that temporal models are particularly effective at preserving relative variation in $R_{max}$ across storms. As shown in Figure 9, all three physics-augmented models reproduce the overall positive correlation between predicted and observed $R_{max}$, although predictions become increasingly dispersed and tend to underestimate the largest observed values.

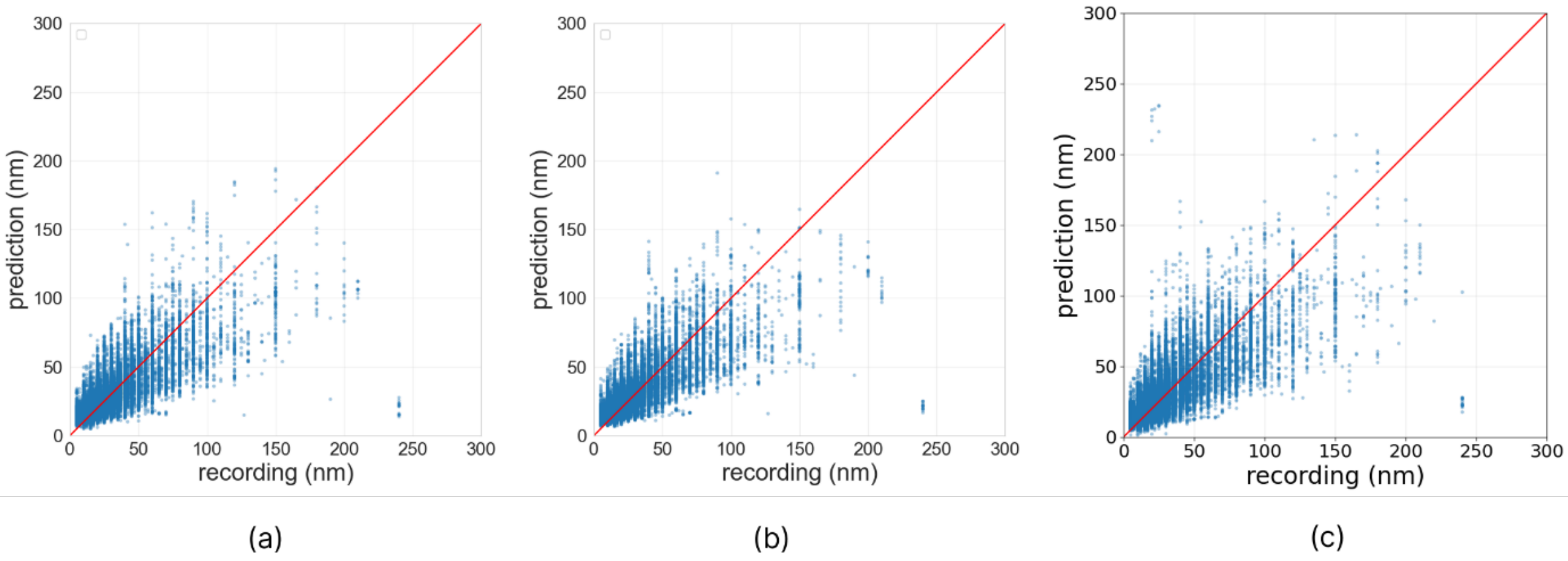

Figure 9. Scatter plot comparing the predicted and observed $R_{max}$ values for (a) LSTM, (b) 1DCNN, and (c) XGBoost using the physics-augmented configuration. The plots displayed include testing data across all folds.

However, as noted above, $R_{34}$ is not widely available in recorded datasets. Thus, models relying on $R_{34}$ as an input variable have limited applicability for imputing $R_{max}$ data for historical storms, despite their higher predictive performance. In the absence of $R_{34}$, the correlation advantage of the temporal models becomes more evident, amplifying their ability to capture variation and patterns in the data. In fact, the reduction in correlation when excluding $R_{34}$ is more substantial for the best performing non-temporal model (XGBoost) than for either temporal model, suggesting that the ability to leverage temporal dependencies within the storm data partially compensates for the missing $R_{34}$ variable in the model's ability to recognize patterns. Across all models, the results suggest that inclusion of temporal information partially compensates for the reduced number of training samples by encoding the evolution of storm structure over multiple time steps rather than a single independent observation.

### 3.2 Error distribution analysis

To further assess model behavior, we visualize predicted $R_{max}$, observed $R_{max}$, and prediction error against storm variables for the LSTM model with the physics-augmented input configuration, which is the best performing temporal model, and for the LSTM model with the input configuration excluding $R_{34}$. Results are shown in Figure 10 and Figure 11. Similar trends are observed across both input configurations. Results show that storms characterized by high $V_{max}$ and low $P_c$ tend to exhibit smaller prediction errors, suggesting that the model imputes $R_{max}$ more reliably for stronger, more structurally consistent storms compared to weaker systems. However, this pattern may also reflect the observed $R_{max}$ values themselves being relatively lower for stronger storms as compared to cases when the $P_c$ is higher, and the $V_{max}$ are lower, resulting in lower errors. The inclusion of the physics-augmented information results in smaller errors associated with less intense storms. It is also possible to observe the effects of the reduced data availability associated with the physics-augmented input configuration, as reflected by the smaller range of observed $R_{max}$ values contained in the testing sets.

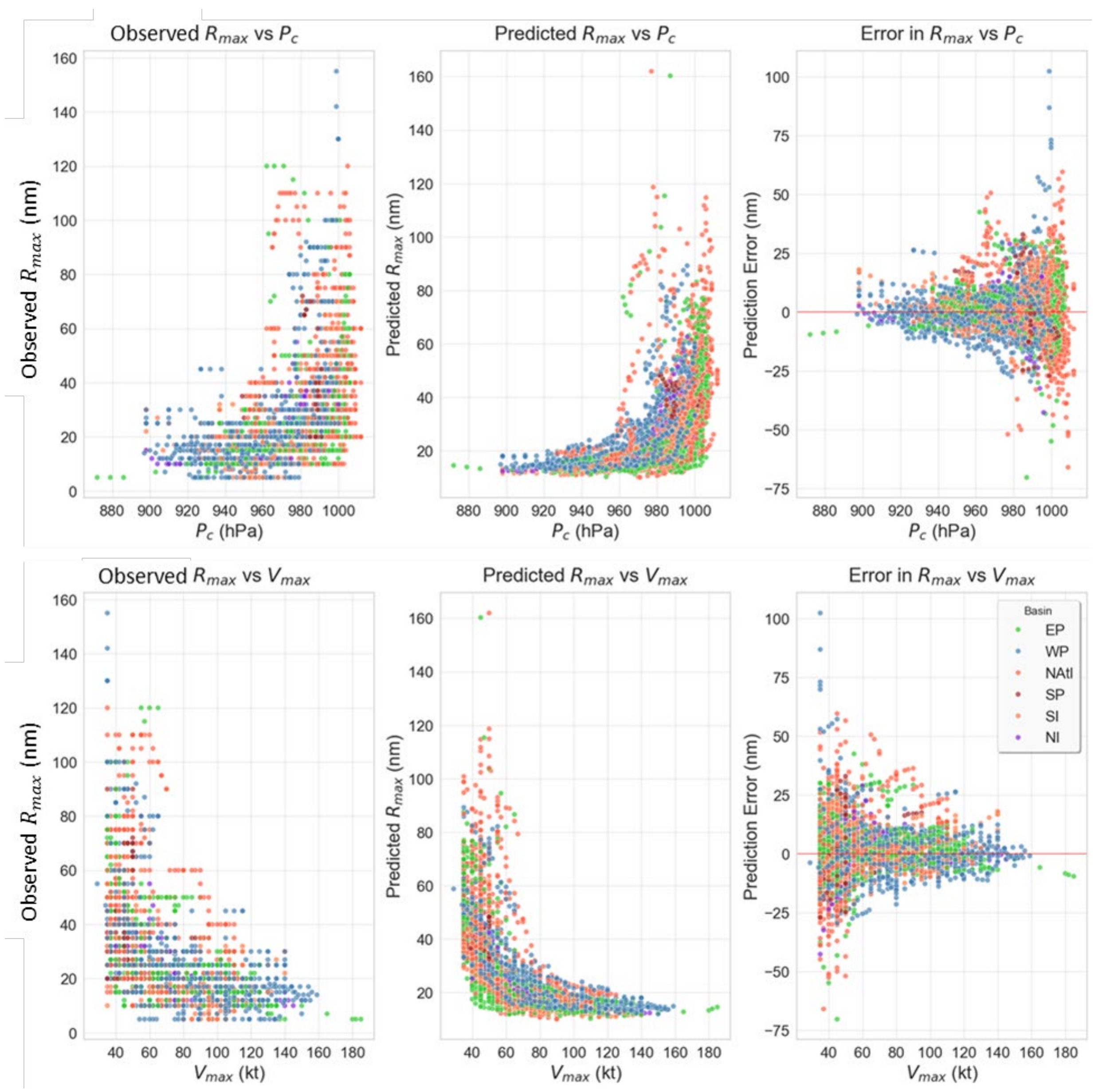

Figure 10. Relationship between observed $R_{\max}$, predicted $R_{\max}$, and prediction error with $P_c$ and $V_{\max}$ for the LSTM model using the physics-augmented input configuration. Points are colored by basin. WP: West North Pacific, EP: East North Pacific, NAtl: North Atlantic, SI: South Indian, SP: South Pacific, NI: North Indian.

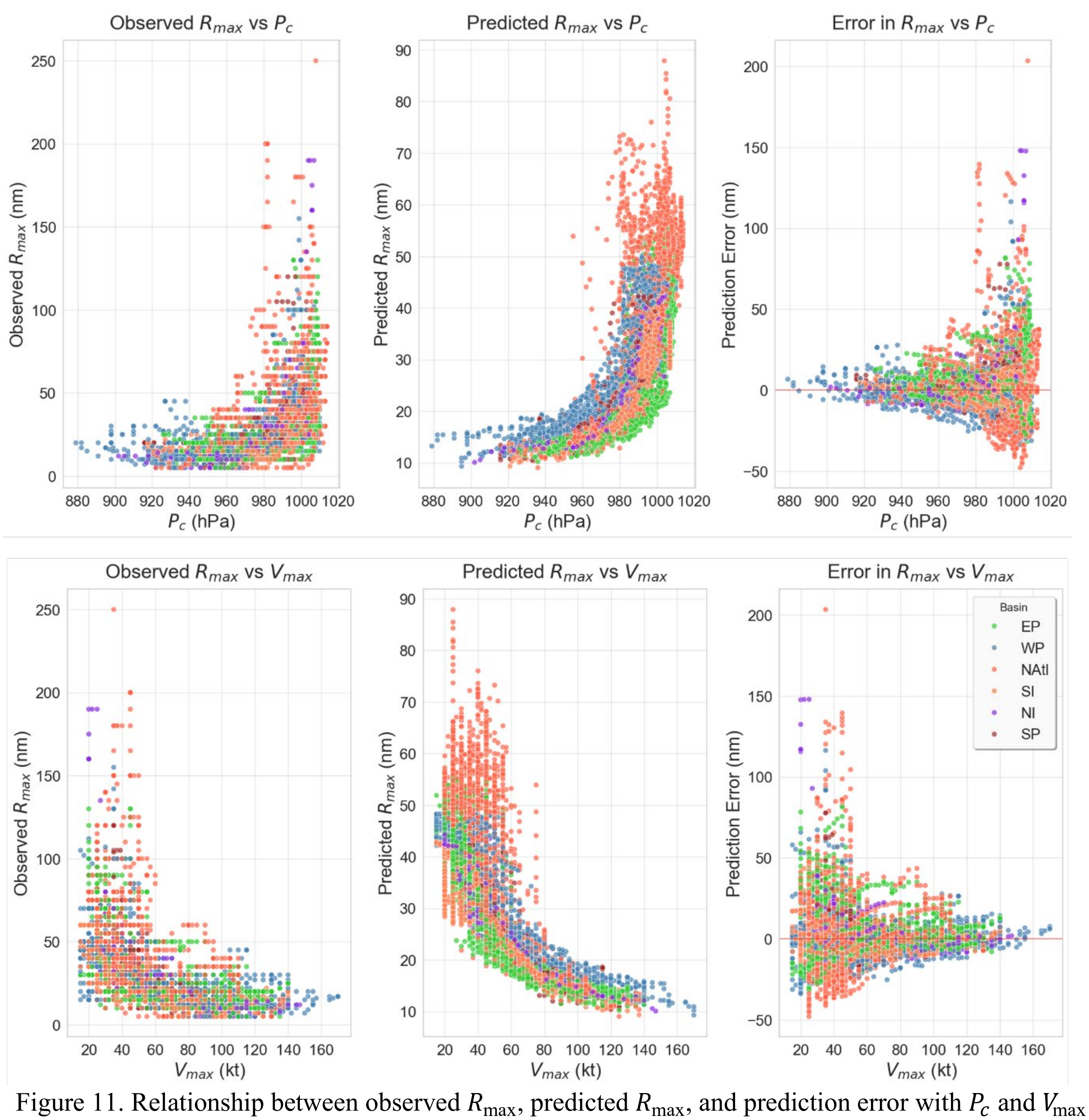

Figure 11. Relationship between observed $R_{\max}$, predicted $R_{\max}$, and prediction error with $P_c$ and $V_{\max}$ for the LSTM model using the input configuration without $R_{34}$. Points are colored by basin. WP: West North Pacific, EP: East North Pacific, NAtl: North Atlantic, SI: South Indian, SP: South Pacific, NI: North Indian.

## 3.3 Transfer learning

Comparative analysis revealed that models pre-trained on synthetic data and subsequently fine-tuned on IBTrACS did not demonstrate improved performance over models trained exclusively on IBTrACS data. To investigate this unexpected result, we conducted a comprehensive distributional analysis comparing the synthetic datasets (RAFT and STORM) with IBTrACS reanalysis data, as shown in Figure 12.

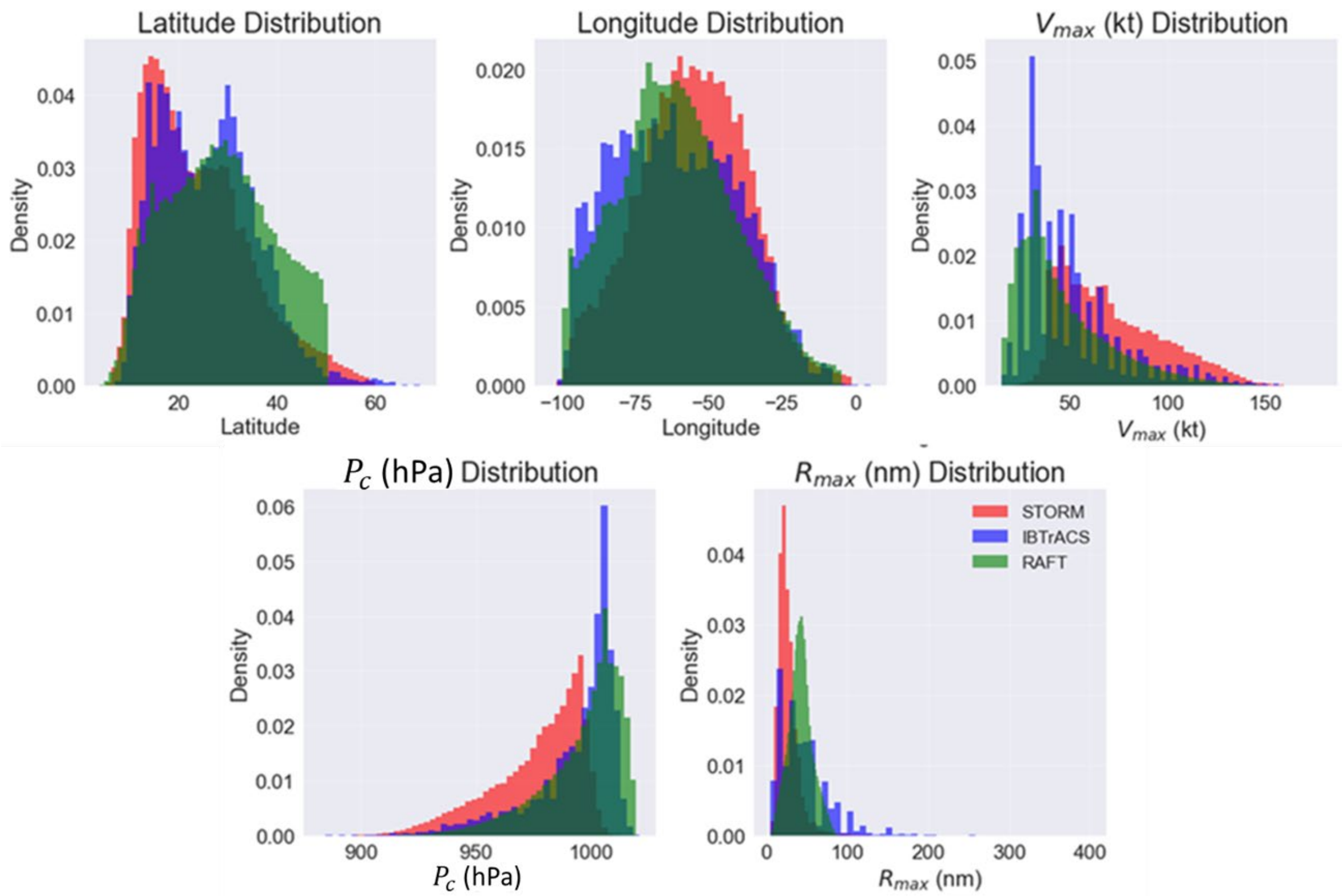

Figure 12. Distribution comparison of key TC parameters across IBTrACS, STORM, and RAFT datasets. The distributions reveal systematic differences in parameter characteristics between real-world observations and synthetic data.

The latitude and longitude distributions indicate broadly comparable geographic coverage across the three datasets, although the relative frequency of observations varies within this coverage. In particular, STORM is more concentrated at lower latitudes, whereas RAFT contains a greater proportion of higher-latitude observations; IBTrACS exhibits features of both distributions. The distributional analysis for other variables reveals several critical differences between synthetic and real-world datasets.

All three datasets show right-skewed distributions for maximum sustained wind speed ($V_{max}$), which is consistent with the greater frequency of lower-intensity TCs relative to high-intensity hurricanes or typhoons. However, IBTrACS demonstrates a much higher concentration of storms with lower maximum wind speeds, with the mode around 30-40 knots. The RAFT dataset follows this trend but is shifted to the left. The STORM dataset is shifted further to the right, indicating that higher wind speeds are more likely.

The minimum central pressure distributions are also consistent with the trends in the $V_{max}$ distributions. IBTrACS shows a much narrower peak at higher pressure values (~1000 hPa). The distributions for the synthetic datasets form relatively lower peaks, with the $P_c$ in the STORM dataset peaking at a lower value (~990 hPa) than in the RAFT dataset (~1000 hPa).

$R_{max}$ shows the most significant differences. Comparative analysis of $V_{max}$ and $P_c$ revealed that IBTrACS data have a higher concentration of storm data points with lower $V_{max}$ and higher $P_c$, which typically result in more weakly developed storms. This is also evident from the $R_{max}$ distribution of the IBTrACS dataset, which has a much longer tail and extends to higher $R_{max}$ values than the synthetic datasets.

Figure 12 suggests that the synthetic datasets are generally centered at lower $R_{max}$ values and exhibit a narrower spread than IBTrACS. This indicates that the synthetic data generation processes may not fully capture the range and variability of $R_{max}$ observed in real storms. This may be due to the parametric or simplified physical models underlying synthetic storm generation, which prioritize computational efficiency and physical consistency over reproducing the full statistical variability of observed storms.

This distributional mismatch likely contributed to the limited effectiveness of the transfer learning approach, as the pre-trained models learned patterns and correlations that were not representative of real-world storm behavior. Specifically, models pre-trained on synthetic data learned to predict smaller $R_{max}$ values on average and did not encounter the wide variability present in observations. When subsequently fine-tuned on IBTrACS data, these models had to "unlearn" synthetic data biases while simultaneously learning real-world patterns, potentially resulting in worse performance than models trained exclusively on IBTrACS.

These findings suggest that for transfer learning to be effective in TC parameter imputation, synthetic datasets must be specifically calibrated to match the statistical distributions of real observations, particularly for the output variable. Future work should investigate methods for post-processing synthetic data to better align with observed distributions or for developing hybrid synthetic-observed training strategies that can leverage the sample size advantages of synthetic data while maintaining distributional fidelity to observations.

## 4 Conclusion

This study evaluates temporal deep learning models, including LSTM networks and 1DCNNs, for imputing missing $R_{max}$ values in TC best-track records. Their performance was compared with non-temporal ML models, including XGBoost and an ANN, as well as the CK22 physics-based parametric model.

The results demonstrate that incorporating $R_{34}$ substantially improves $R_{max}$ imputation across all model types. Among the evaluated models, the physics-augmented LSTM achieved the highest average correlation and the lowest average RMSE. More broadly, the temporal models consistently achieved higher average correlations than the non-temporal models despite being trained on approximately an order of magnitude fewer samples. This finding indicates that temporal models can effectively use information on storm evolution to preserve variations in $R_{max}$ across storms.

When $R_{34}$ was excluded, the performance of all models declined. However, the temporal models retained a clearer performance advantage over the non-temporal models, suggesting that temporal dependencies partially compensate for the absence of this important storm size predictor. This result is particularly relevant for historical $R_{max}$ imputation because $R_{34}$ is frequently unavailable in earlier best-track records. Nevertheless, constructing temporally continuous input blocks substantially reduces the number of usable training samples, highlighting a trade-off between representing storm evolution and maximizing data availability.

Transfer learning using the RAFT and STORM synthetic TC datasets did not improve model performance relative to training exclusively on IBTrACS data. Distributional comparisons showed that the synthetic datasets contain systematically smaller and less variable $R_{max}$ values than the observational data. Consequently, pre-trained models learned relationships that were not fully representative of observed storm behavior. These findings demonstrate that the size of a pre-training dataset alone is insufficient to support effective transfer learning. The synthetic data must also reproduce the statistical distribution and variability of the target observational dataset, particularly for the predicted variable.

Future work should investigate temporal architectures that dynamically incorporate available observed $R_{max}$ values within otherwise incomplete storm sequences. Additional efforts should focus on distribution-alignment or domain-adaptation methods for synthetic data, hybrid synthetic-observational training strategies, and evaluation frameworks that separate hyperparameter selection from final model assessment. These developments could improve the generalizability and reliability of deep learning methods for reconstructing incomplete TC records and supporting probabilistic coastal hazard assessments.

## 5 Statements and Declarations

### Acknowledgments

This work builds upon the collaboration in Liu et al. (2024b). The authors gratefully acknowledge previous collaborations with Dr. Meredith L. Carr, Dr. Norberto C. Nadal-Caraballo, Madison C. Yawn, and Dr. Alexandros A. Taflanidis.

### Funding

This research did not receive any specific grant from funding agencies in the public, commercial, or not-for-profit sectors.

### Data Availability

The observational data used in this study were obtained from the International Best Track Archive for Climate Stewardship (IBTrACS) (https://www.ncei.noaa.gov/products/international-best-track-archive). Synthetic tropical cyclone data from the North Atlantic synthetic tropical cyclone track, intensity, and rainfall dataset (RAFT) (https://data.pnnl.gov/group/nodes/dataset/33569) and Synthetic Tropical cyclOne geneRation Model (STORM) (https://research.vu.nl/en/datasets/storm-ibtracs-present-climate-synthetic-tropical-cyclone-tracks/) datasets were also used to investigate transfer learning.

### Competing Interests

The authors have no relevant financial or non-financial interests to disclose.

### Author Contributions

SA: Conceptualization, Methodology, Software, Investigation, Formal analysis, Data Curation, Visualization, Writing-Original Draft, Writing-Review & Editing. NH: Conceptualization, Methodology, Software, Investigation, Formal analysis, Data Curation, Visualization, Writing-Original Draft, Writing-Review & Editing. ZL: Conceptualization, Methodology, Software, Investigation, Formal analysis, Data Curation, Visualization, Writing-Original Draft, Writing-Review & Editing, Supervision. MB: Conceptualization, Data Curation, Methodology, Visualization, Supervision, Writing-Review & Editing.

**Appendix A –** Earth-centered, Earth-fixed Coordinate Transformation

Latitude and longitude were converted to ECEF Cartesian coordinates using the WGS84 reference ellipsoid. The ECEF coordinates $X$, $Y$, and $Z$ are computed as:

$$X = (N + h)\cos(\mathrm{lat})\cos(\mathrm{lon}) \tag{5}$$

$$\mathrm{Y} = (N + h)\cos(\mathrm{lat})\sin(\mathrm{lon}) \tag{6}$$

$$Z = [(1 - e^2)N + h]\sin(\mathrm{lat}) \tag{7}$$

where

$$N = \frac{a}{\sqrt{1 - e^2\sin^2(\mathrm{lat})}} \tag{8}$$

Here, lat is the geodetic latitude in radians, lon is the longitude in radians, $h$ is the ellipsoidal height, $a$ is the semi-major axis of the WGS84 ellipsoid, and $e^2$ is the first eccentricity squared. The first eccentricity squared is defined as:

$$e^2 = f_{\mathrm{WGS84}}(2 - f_{\mathrm{WGS84}}) \tag{9}$$

where $f_{\mathrm{WGS84}} = 1/298.26$ is the WGS84 flattening factor. Because the storm track data do not include elevation, $h$ is set to zero in this study.